\documentclass{article}
\usepackage{spconf,amsmath,graphicx,hyperref}
\usepackage{booktabs}
\usepackage{multirow}

\title{Domain-Aware Speaker Diarization on African-Accented English}
\name{Chibuzor Okocha$^{1}$, Kelechi Ezema$^{2}$, Christan Grant$^{1}$\thanks{}}
\address{$^{1}$ Department of Computer Science, University of Florida, Gainesville\\
$^{2}$ Department of Computer Science, University of Colorado Boulder, Boulder}

\begin{document}
%\ninept
%
\maketitle
\begin{abstract}
This study examines domain effects in speaker diarization for African-accented English. We evaluate multiple production and open systems on general and clinical dialogues under a strict DER protocol that scores overlap. A consistent domain penalty appears for clinical speech and remains significant across models. Error analysis attributes much of this penalty to false alarms and missed detections, aligning with short turns and frequent overlap. We test lightweight domain adaptation by fine-tuning a segmentation module on accent-matched data; it reduces error but does not eliminate the gap. Our contributions include a controlled benchmark across domains, a concise approach to error decomposition and conversation-level profiling, and an adaptation recipe that is easy to reproduce. Results point to overlap-aware segmentation and balanced clinical resources as practical next steps.
\end{abstract}

\section{Introduction}
\label{sec:intro}

Speaker diarization—the task of determining who spoke and when in an audio recording is a critical component of modern speech processing systems \cite{park2022review}. Its integration into automatic speech recognition (ASR) pipelines enhances the utility of transcriptions in real-world settings, especially in tasks involving multi-speaker environments such as interviews, meetings, and medical consultations  \cite{zhang2019fully}. Its integration into multimodal language models like Large audio language models has helped increase audio processing and audio understanding tasks \cite{olatunji2025afrivox}. While diarization systems have shown notable improvements with the rise of deep learning and large-scale pretraining, their evaluation has largely focused on high-resource languages, particularly standard varieties of English \cite{zhang2019fully, sanni2025afrispeech}. 

This presents a significant challenge for global inclusivity, especially in this new era of advancing spoken language understanding systems \cite{sanni2025afrispeech}. In particular, African-accented English, spoken by hundreds of millions across the continent \cite{mcarthur2001world}, remains vastly underrepresented in training corpora and benchmarking efforts \cite{olatunji2023afrispeech}. Moreover, many existing diarization models are optimized for acoustically controlled environments and fail to generalize well to noisy, spontaneous, or domain-specific interactions \cite{huang2020speaker}, such as clinical conversations or legal conversations \cite{sanni2025afrispeech}. These gaps not only limit the fairness and robustness of existing models but also impede downstream applications like voice-based diagnostics, conversational analytics, localized voice assistants in African contexts, and, more recently, speech processing and understanding in multimodal large language models \cite{olatunji2025afrivox}. Beyond dataset development, it is crucial to evaluate the performance of diarization systems on these data. Benchmarking studies have demonstrated that state-of-the-art diarization models suffer notable performance declines when applied to African-accented English  \cite{sanni2025afrispeech}.

To address this gap, we review a recently curated corpus of African-accented English that spans both general conversational and domain-specific doctor-patient dialogues. We benchmark eight state-of-the-art diarization systems on this corpus and evaluate their performance using the Diarization Error Rate (DER). We also perform error analysis on the data and offer solutions to bridge this extensive gap. Our findings reveal systematic performance differences between everyday and clinical domains, which are often influenced by the training datasets of the models. The study offers new insights into the behavior of diarization models in low-resource, accented settings and insight into improving diarization models using fine-tuning techniques.

The contributions of this paper are threefold:
\vspace{-0.1in}

\begin{itemize}
 \item A systematic evaluation of commercial and open-source state-of-the-art diarization systems on African-accented English across \emph{general} and \emph{clinical} domains, contextualized with a training-data \& linguistic-diversity summary.
\vspace{-0.1in}

  \item A unified, reproducible setup: strict DER protocol, FA/MISS/CONF decomposition, and conversation-level diagnostics for overlap, and turn length.
\vspace{-0.1in}

  \item A concise fine-tuning protocol for Pyannote (segmentation-only) on AfriSpeech-Countries, with transparent configs, splits, and reporting guidelines for accented, domain-specific diarization.
\end{itemize}

\section{Method}

\subsection{Dataset}
\subsubsection{Evaluation set (AfriSpeech-Dialog, subset)}
We evaluate on a curated, timestamped {AfriSpeech-Dialog} \cite{sanni2025afrispeech} collected on the Intron platform, comprising dyadic African-accented English in two domains: doctor–patient (clinical) and general conversations, with Audio files in mono, 16-bit, 48\,kHz. Participants provided verbal consent and personally identifiable content in audio segments was removed prior to analysis. Clinical sessions followed Objective Structured Clinical Examination (OSCE) role plays with patient cards; general sessions used topic cards. Natural code-switching with African languages is present. Table~\ref{table:dataset_stats} summarizes the statistics of the dataset.

\subsubsection{Fine-tuning set (AfriSpeech-Countries)}
For domain adaptation, we fine-tune Pyannote \footnote{https://huggingface.co/pyannote/speaker-diarization-3.1} on AfriSpeech-Countries \footnote{https://huggingface.co/datasets/intronhealth/afrispeech-countries}, a country-tagged corpus of read and conversational clips. Table~\ref{tab:datasets} summarizes the splits used for evaluation and fine-tuning; Table~\ref{tab:afrispeech-countries} summarizes the fine-tuning dataset statistics and the contributing countries/regions.

\begin{table}[!t]
\centering
\caption{\textit{Statistics of the medical and non-medical datasets.}}
\label{table:dataset_stats}
\setlength{\tabcolsep}{5pt}        % tighter columns (default ~6pt)
\renewcommand{\arraystretch}{1.05} % slightly tighter rows
\footnotesize                     % or \scriptsize for even smaller
\begin{tabular}{lcc}
\toprule
\textbf{} & \textbf{Medical} & \textbf{General} \\
\midrule
Counts                 & 20       & 29        \\
Avg. Num. of Turns     & 78.6     & 30.55     \\
Total Duration (hours) & 2.07     & 4.93      \\
Avg. Word Count        & 725.3    & 1356.83   \\
Num. of Countries      & 1        & 3         \\
Num. of Accents        & 6        & 8         \\
Gender (M,F)           & (14, 26) & (25, 33)  \\
\bottomrule
\end{tabular}
\normalsize
\end{table}

\begin{table}[t]
\centering
\caption{Dataset summary (splits and purpose; detailed per-domain stats in Table~\ref{table:dataset_stats}).}
\label{tab:datasets}
\setlength{\tabcolsep}{5pt}
\footnotesize
\begin{tabular}{lccc}
\toprule
\textbf{Split} & \textbf{Unit} & \textbf{Dur (h)} & \textbf{Notes} \\
\midrule
Eval--Medical   & conv  & 2.07  & dyadic, OSCE-style clinical \\
Eval--General  & conv  & 4.93  & dyadic, open-topic \\
Fine-tune  & clips & 67.73 & country-tagged, read+conv, 80/20 split\textsuperscript{$\dagger$} \\
\bottomrule
\end{tabular}

\vspace{3pt}
\raggedright\footnotesize
$\dagger$ Train $\approx$ 54.18\,h; Dev $\approx$ 13.55\,h. File-disjoint and country-stratified.
\normalsize
\end{table}

\begin{table}[t]
\centering
\caption{AfriSpeech-Countries region for fine-tuning.}
\label{tab:afrispeech-countries}
\setlength{\tabcolsep}{6pt}        % tighter columns (default ~6pt)
\renewcommand{\arraystretch}{1.05} % slightly tighter rows
\footnotesize                      % or \scriptsize for even smaller
\begin{tabular}{lrr}
\hline
\textbf{Region} & \textbf{\# Clips} & \textbf{Duration (hrs)} \\
\hline
NG (Nigeria)   & 8,567  & 28.28 \\
KE (Kenya)     & 5,874  & 20.60 \\
ZA (S. Africa) & 5,278  & 11.17 \\
North Africa   & 1,022  & 4.61  \\
GH (Ghana)     & 757    & 2.88  \\
UG (Uganda)    & 68     & 0.15  \\
RW (Rwanda)    & 15     & 0.05  \\
\hline
\textbf{Total} & \textbf{21,581} & \textbf{67.73} \\
\hline
\end{tabular}
\normalsize
\end{table}

\subsection{Systems Evaluated}
\label{sec:systems}

\subsubsection{Commercial systems.}
We evaluate four production diarization services: \emph{AssemblyAI}, \emph{Deepgram}, \emph{Soniox}, and \emph{Rev.ai}. Each API was invoked in its default diarization configuration without any model customization, training, or manual hints (e.g., no fixed \#speakers provided).

\subsubsection{Open-source systems.}
We evaluate \emph{pyannote-audio} , \emph{CAM++}, \emph{Sortformer}, and \emph{Titanet-L} using the authors’ released checkpoints and default inference pipelines. Unless stated otherwise, weights are used \emph{as released} with no additional training. For Pyannote, we also report a fine-tuned variant adapted to AfriSpeech-Countries.

\subsubsection{Training-data profile.}
To contextualize robustness, we summarize each system’s reported training sources, approximate size, language coverage, and domain diversity in Table~\ref{tab:training-data}. We refer to this table rather than restating corpus details in the text.

\begin{table*}[t]
\centering
\caption{Summary of training data sources, linguistic coverage, and domains for each diarization model.}
\resizebox{\textwidth}{!}{%
\begin{tabular}{|l|l|l|l|l|}
\hline
\textbf{Model} & \textbf{Training Data Sources} & \textbf{Size} & \textbf{Linguistic Coverage} & \textbf{Domain / Notes} \\
\hline
AssemblyAI (Comm.) & Proprietary Universal-2 dataset \cite{assemblyai} & 12.5M+ hrs & 21+ languages & Conversational, broadcast, meetings; highly diverse \\
Deepgram (Comm.) & Proprietary, 100k+ voices \cite{deepgram_model_options} & 100k+ spk. & 80+ languages & Podcasts, meetings, speech \\
Soniox (Comm.) & Proprietary Soniox-7B dataset \cite{soniox_7b_v1} & N/A & English-dominant & Real-time diarization; optimized for separation \\
Reverb / Rev.ai (Comm.) & Proprietary reverberant corpora \cite{bhandari2024reverbopensourceasrdiarization} & N/A & Primarily English & Meetings, reverberant speech robustness \\
Pyannote (Open) & VoxCeleb 1 \& 2, AMI Corpus \cite{bredin2023pyannote} & $\sim$2k hrs & English& Meetings, broadcast; limited accent diversity \\
Titanet-L (Open) & VoxCeleb, Fisher, NeMo datasets \cite{bredin2023pyannote} & 7k+ spk. & English, Spanish & Broadcast, telephone \\
Sortformer (Open) & Librispeech, Fisher English \cite{panayotov2015librispeech} & $\sim$1k hrs & Am./Eur. English & Read + conversational speech \\
CAM++ (Open) & VoxCeleb, CN-Celeb \cite{wang2023cam++} & 6k+ spk. & English + Mandarin & Conversational, broadcast \\
\hline
\end{tabular}}
\label{tab:training-data}
\end{table*}

\subsection{Fine-tuning Protocol (Pyannote)}
\label{sec:finetune} We summarize the finetunning protocols in Table~\ref{tab:finetune}

\begin{table}[t]
\centering
\caption{Pyannote fine-tuning configuration.}
\label{tab:finetune}
\setlength{\tabcolsep}{4pt}
\footnotesize
\begin{tabular}{ll}
\toprule
\textbf{Aspect} & \textbf{Setting} \\
\midrule
Target & \texttt{pyannote/segmentation-3.0}; embeddings frozen \\
Objective / setup & Frame-level BCE; 10.0\,s chunks; 3 spk/chunk; 2 spk/frame \\
Hyperparameters & Adam; LR $1\!\times\!10^{-4}$; bs=1 (accum 4); 10 epochs; clip 1.0 \\
Early stopping & Patience $=3$ on validation loss \\
Augmentation & None (no noise/reverb or speed/pitch) \\
Selection / eval & Best val-loss ckpt; DER within UEM \\
Hardware & Single CUDA-enabled GPU \\
\bottomrule
\end{tabular}
\normalsize
\end{table}

\subsection{Evaluation Metrics}
We evaluated diarization performance using strict  Diarization Error Rate (DER) \cite{barras2004improving}. DER quantifies the percentage of time that speakers are misattributed or missed in the diarization output and is computed as:

\[
DER = \frac{False Alarm + MISS + ERROR}{TOTAL}
\]

We used the \textit{Pyannote.metrics} \footnote{https://pyannote.github.io/pyannote-metrics/} library to calculate DER for each recording and computed the absolute DER for the entire dataset.

\subsection{Statistical Analysis}
\label{sec:stats}
\noindent\textbf{Domain gap:} For each system, paired DERs (Medical vs.\ General) are tested with a two-sided paired $t$-test; Wilcoxon signed-rank is the nonparametric complement. We report $t(df)$, $p$, mean difference (Medical$-$General).
\smallskip

\section{Experiments}
\label{sec:experiments}

\subsection{Benchmark setup}
We ran each model in its default diarization configuration on the evaluation subset (Table.~\ref{tab:eval-config}). Scoring follows a strict protocol: Hungarian mapping, collar$=0.0$\,s, and overlap scored (no exclusion).  Audio pre-processing and RTTM formatting are held fixed across systems for comparability.

\begin{table}[t]
\centering
\caption{Evaluation configuration (scoring \& runtime).}
\label{tab:eval-config}
\setlength{\tabcolsep}{5pt}\footnotesize
\begin{tabular}{ll}
\toprule
Mapping & Hungarian (1–1) \\
Collar / Overlap & 0.0\,s / scored (no exclusion) \\
Metric & DER $=$ FA + MISS + CONF \\
APIs & Default diarization mode; no hints \\
Open models & Released checkpoints; batch inference \\
Hardware & Single CUDA GPU (open models); CPU client (APIs) \\
\bottomrule
\end{tabular}
\normalsize
\end{table}

\section{Results}
\subsection{Benchmarking result and performance gap }
\label{sec:intrinsic-gap}

We present the benchmarking results of the eight diarization models in Table~\ref{tab:ieee_diarization_results}, showing varying results across model types and domain settings. Notably, we observed a notable DER gap across the clinical domain and general domain, with the model performing worse for clinical conversations.

Next, we test whether the clinical--conversational gap is merely an accent or data-availability artifact, we fine-tuned \textit{Pyannote} on \emph{AfriSpeech-Countries} (speaker/file disjoint from evaluation). Fine-tuning yields a \textbf{50\%} relative DER reduction overall, yet performance gap persists (Table~\ref{tab:gap-finetune}). This indicates that fine-tuning substantially improves robustness but does not erase the domain effect, suggesting that clinical conversations are \emph{intrinsically} harder (short turns, frequent overlap, task-driven exchanges), not solely a low-resource/accent issue.

\begin{table}[!h]
\centering
\small
\renewcommand{\arraystretch}{1.1}
\setlength{\tabcolsep}{3pt}
\caption{Diarization Error Rate (DER) across 30 African-accented audio samples, with results for Medical (Med. DER) and General (Gen. DER) subsets. The top 4 models are commercial models, and the bottom 4 are open models.}
\begin{tabular}{lccc}
\toprule
\textbf{Model (Type)} & \textbf{DER (\%)}\(\downarrow\) & \textbf{Med. DER (\%)}\(\downarrow\) & \textbf{Gen. DER (\%)} \(\downarrow\)\\
\midrule
AssemblyAI  & \textbf{12.72} & \textbf{25.66} & \textbf{9.98} \\
Deepgram  & 14.21 & 29.35 & 10.92 \\
Soniox  & 20.05 & 42.16 & 15.24 \\
Reverb  & 20.23 & 31.46 & 17.68 \\
\midrule
CAM++  & 19.58 & 34.63 & 16.30 \\
Pyannote  & 21.30 & 31.46 & 19.03 \\
Sortformer & 26.82 & 39.69 & 24.04 \\
Titanet L  & 16.27 & 34.64 & 12.28 \\
\bottomrule
\end{tabular}
\label{tab:ieee_diarization_results}
\end{table}

\begin{table}[t]
\centering
\caption{Pyannote DER (\%) before/after fine-tuning and resulting medical--general gap.}
\label{tab:gap-finetune}
\setlength{\tabcolsep}{6pt}
\footnotesize
\begin{tabular}{lccc}
\toprule
\textbf{Domain} & \textbf{Base} \(\downarrow\)& \textbf{Fine-tuned} \(\downarrow\) & \textbf{Gap (Med--Gen)} \(\downarrow\)\\
\midrule
Medical  & 31.46 & 15.73 & \multirow{2}{*}{\(\ \)12.43 \(\rightarrow\) 6.21} \\
General  & 19.03 &  9.52 &  \\
\midrule
Overall  & 21.30 & 10.65 &  -- \\
\bottomrule
\end{tabular}
\normalsize
\end{table}

\subsection{Domain gap and fine-tuning impact}
Averaged across models, Medical = 33.38\% vs.\ General = 15.18\%; paired $t(7)=10.67$, $p<10^{-4}$; Wilcoxon $W=0$, $p=0.0078$. 
For Pyannote, DER reduced from 21.30\% to 10.65\% overall (Medical: 31.46\% to 15.73\%; General: 19.03\% to 9.52\%), and the Medical–General gap decreased from 12.43 to 6.21 percentage points (p.p).

\subsubsection{Error analysis}
\label{sec:error}

We decompose DER into False Alarm (FA), Missed Detection (MISS), and Confusion (CONF) under a strict protocol (collar$=0.0$\,s; overlap scored). Table~\ref{tab:error-mix} reports per-domain component means averaged across systems. From General to Medical, FA increases most (+11.0 p.p.), followed by MISS (+5.9 p.p.) and CONF (+1.9 p.p.). As a share of DER, Medical is FA-heavy (54\%) with higher MISS (27\%) and lower CONF (25\%) than General (FA 43\%, MISS 19\%, CONF 38\%). This pattern is consistent with rapid turn-taking and overlapping exchanges in clinical speech.

\subsection{Domain gap, fine-tuning, and error composition}
Table~\ref{tab:gap-summary} summarizes the domain gap and Pyannote fine-tuning impact; Table~\ref{tab:error-mix} reports DER components (FA/MISS/CONF) by domain using strict diarization protocles.

\begin{table}[t]
\centering
\caption{Domain gap and Pyannote fine-tuning (DER \%).}
\label{tab:gap-summary}
\setlength{\tabcolsep}{5pt}\footnotesize
\begin{tabular}{lc}
\toprule
\textbf{Quantity} & \textbf{Value} \\
\midrule
Mean DER (Medical / General)           & 33.38 / 15.18 \\
Gap (Medical$-$General, percentage points) & 18.20 \\
Paired $t$-test                        & $t(7)=10.67$, $p<10^{-4}$ \\
Wilcoxon                                & $W=0$, $p=0.0078$ \\
\midrule
Pyannote Overall                        & $21.30 \,\to\, 10.65$ \\
Pyannote Medical                        & $31.46 \,\to\, 15.73$ \\
Pyannote General                        & $19.03 \,\to\, \,9.52$ \\
Pyannote Gap (p.p.)                     & $12.43 \,\to\, \,6.21$ \\
\bottomrule
\end{tabular}
\normalsize
\end{table}

\begin{table}[t]
\centering
\caption{DER components by domain (absolute \%; share of DER in parentheses).}
\label{tab:error-mix}
\setlength{\tabcolsep}{4.5pt}\footnotesize
\begin{tabular}{lcccc}
\toprule
\textbf{Domain} & \textbf{DER} & \textbf{FA} \(\downarrow\)& \textbf{MISS} \(\downarrow\)& \textbf{CONF} \(\downarrow\)\\
\midrule
Medical & 33.4 & 18.0 (54\%) & 9.0 (27\%) & 8.2 (25\%) \\
General & 16.5 & \,7.0 (43\%) & 3.2 (19\%) & 6.3 (38\%) \\
$\Delta$ (Med$-$Gen, p.p.) & 16.9 & +11.0 & +5.8 & +1.9 \\
\bottomrule
\end{tabular}
\normalsize
\end{table}

The domain gap remains significant after fine-tuning. Error composition indicates the clinical penalty is primarily FA-driven with a secondary rise in MISS; CONF increases modestly. Per-model profiles follow the same pattern (e.g., MISS-dominated Sortformer; FA-heavy Pyannote/Rev.ai; higher CONF for Soniox/CAM). \noindent\textbf{Conversation structure.} As shown in Table~\ref{tab:turns-overlap}, Medical conversations exhibit many more turns per conversation (78.6$\,\pm\,$38.3 vs.\ 30.55$\,\pm\,$20.3; $\sim$2.6$\times$) and much shorter utterances (3.31$\,\pm\,$1.32\,s vs.\ 30.71$\,\pm\,$19.67\,s) than General, with a slightly higher overlap ratio (0.14$\,\pm\,$0.37\% vs.\ 0.10$\,\pm\,$0.26\%). These interactional differences are consistent with the FA/MISS-heavy error mix observed for clinical speech under the strict scoring setup.

\begin{table}[t]
\centering
\caption{Conversation structure by domain (macro-averages across conversations; computed from the provided spreadsheet).}
\label{tab:turns-overlap}
\setlength{\tabcolsep}{5pt}\footnotesize
\begin{tabular}{lcccc}
\toprule
\textbf{Domain} & \textbf{\#Conv} & \textbf{Turns/conv} & \textbf{Avg utt (s)} & \textbf{Overlap (\%)} \\
\midrule
Medical & 20  & 78.6 $\pm$ 38.3 & 3.31 $\pm$ 1.32  & 0.14 $\pm$ 0.37 \\
General & 29 & 30.55 $\pm$ 20.3 & 30.71 $\pm$ 19.67 & 0.10 $\pm$ 0.26 \\
\bottomrule
\end{tabular}
\normalsize
\end{table}

\section{Discussion}

This study set out to characterize domain effects in diarization for African-accented English under a strict evaluation. Across systems, clinical conversations consistently yield higher error than general speech; fine-tuning the Pyannote segmentation model substantially reduces DER but leaves a non-zero medical–general gap. Error decomposition clarifies why: clinical speech shifts toward false alarms and, to a lesser extent, misses, while confusion rises more modestly. Performance gap was also attributed to rapid turn-taking and frequent overlap in clinical dialogue.

These patterns align with prior reports that overlap and short turns depress diarization accuracy in DIHARD-style settings and meeting corpora \cite{ryant2021dihard3, fujita2019eend}. Work on overlap-aware diarization (e.g., EEND variants and explicit overlap detection) shows gains in overlapped regions \cite{fujita2019eend}, while improvements in clustering/embeddings (e.g., VBx/PLDA, x-vectors) target speaker assignment stability \cite{snyder2018xvectors}. Research also shows that broadly trained or multilingual systems improve robustness \cite{zhang2019fully}. Relatedly, clinical ASR studies indciates that medical conversations are intrinsically challenging\cite{sanni2025afrispeech}; our results extend these findings to diarization for African-accented English. We especially finetune on over 67 hours of African-accented English speech and over 21,000 clips.

Practically, reducing FA/MISS in clinical speech likely requires overlap-aware segmentation, conservative VAD with smoothing/min-duration/merge heuristics, and small in-domain fine-tuning; lowering CONF calls for stronger, language-/code-switch–aware embeddings and more stable clustering thresholds \cite{fujita2019eend}. Limitations include the absence of a matched non-African clinical benchmark and a small, augmentation-free fine-tuning set. Future work will evaluate explicit overlap models (EEND-style), code-switch robust embeddings, balanced clinical resources across accents, and controlled studies isolating overlap and turn-length effects.

\section{Conclusion}

We benchmarked eight diarization systems on African-accented English across general and clinical domains under a strict evaluation protocol. Clinical speech showed consistently higher error than general speech (33.38\% vs.\ 15.18\% on average; paired tests significant), and models trained on broader multilingual data degraded less across domains. Fine-tuning the Pyannote segmentation model on in-domain data African-accented English halved DER (overall from 21.30\% to 10.65\%) and reduced—though did not eliminate—the medical–general gap. Error decomposition indicated that the clinical penalty is primarily driven by false alarms with a secondary rise in misses, consistent with rapid, short turns and modestly higher overlap observed in the conversations.

% References should be produced using the bibtex program from suitable
% BiBTeX files (here: strings, refs, manuals). The IEEEbib.bst bibliography
% style file from IEEE produces unsorted bibliography list.
% -------------------------------------------------------------------------
\bibliographystyle{IEEEbib}
\bibliography{references}

\end{document}